\documentclass[10pt,twocolumn,letterpaper]{article}

\usepackage{cvpr}              
\usepackage[accsupp]{axessibility}
\usepackage{graphicx}
\usepackage{mathrsfs}
\usepackage{multirow}
\usepackage{amsmath}
\usepackage{amssymb}
\usepackage{booktabs}
\usepackage[dvipsnames]{xcolor}
\renewcommand\footnotemark{}
\definecolor{cvprblue}{rgb}{0.21,0.49,0.74}
\usepackage[pagebackref,breaklinks,colorlinks,citecolor=cvprblue]{hyperref}

\usepackage[capitalize]{cleveref}
\crefname{section}{Sec.}{Secs.}
\Crefname{section}{Section}{Sections}
\Crefname{table}{Table}{Tables}
\crefname{table}{Tab.}{Tabs.}

\definecolor{ben}{rgb}{0.9,0.,0.5}

\def\cvprPaperID{3797} 
\def\confName{CVPR}
\def\confYear{2024}

\begin{document}

\title{SecondPose: SE(3)-Consistent Dual-Stream Feature Fusion\\for Category-Level Pose Estimation}

\author{Yamei Chen$^{1,*}$, Yan Di$^{1,*}$, Guangyao Zhai$^{1,2,\dag}$, Fabian Manhardt$^{3}$, Chenyangguang Zhang$^{4}$,\\ Ruida Zhang$^{4}$,
Federico Tombari$^{1,3}$,
Nassir Navab$^{1}$ and Benjamin Busam$^{1,2}$\\[0.5em]
\textsuperscript{1} Technical University of Munich\quad
\textsuperscript{2} Munich Center for Machine Learning\\
\textsuperscript{3} Google\quad
\textsuperscript{4} Tsinghua University \\[0.5em]
{\url{https://github.com/NOrangeeroli/SecondPose.git}}
\\
\thanks{$^*$ Equal contributions.}
\thanks{$^\dag$ Corresponding author (e-mail: {\tt guangyao.zhai@tum.de}).}
}
\maketitle

\begin{abstract}
Category-level object pose estimation, aiming to predict the 6D pose and 3D size of objects from known categories, typically struggles with large intra-class shape variation.
Existing works utilizing mean shapes often fall short of capturing this variation.
To address this issue, we present SecondPose, a novel approach integrating object-specific geometric features with semantic category priors from DINOv2.
Leveraging the advantage of DINOv2 in providing SE(3)-consistent semantic features, we hierarchically extract two types of SE(3)-invariant geometric features to further encapsulate local-to-global object-specific information.
These geometric features are then point-aligned with DINOv2 features to establish a consistent object representation under SE(3) transformations, facilitating the mapping from camera space to the pre-defined canonical space, thus further enhancing pose estimation.
Extensive experiments on NOCS-REAL275 demonstrate that SecondPose achieves a 12.4\% leap forward over the state-of-the-art.
Moreover, on a more complex dataset HouseCat6D which provides photometrically challenging objects, SecondPose still surpasses other competitors by a large margin.
\end{abstract}

\section{Introduction}
\begin{figure}[t]
    \centering
    \includegraphics[width=0.45\textwidth]{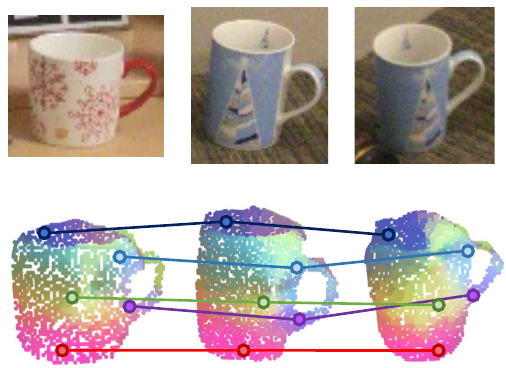}
    \caption{\textbf{Categorical SE(3)-consistent features.} We visualize our fused features by PCA. Colored points highlight the most corresponding parts, where our proposed feature achieves consistent alignment cross instances (left vs. middle) and maintains consistency on the same instance of different poses (middle vs. right).}
    \label{fig:teaser}
\vspace{-0.5cm}
\end{figure}

Category-level pose estimation involves estimating the complete 9 degrees-of-freedom (DoF) object pose, encompassing 3D rotation, 3D translation, and 3D metric size, for arbitrary objects within a known set of categories. This task has garnered significant research interest due to its essential role in various applications, including the AR/VR industry~\cite{marchand2015pose, tan20176d, Tjaden_2017_ICCV, zhang2023ddf}, robotics~\cite{zhai2022monograspnet, zhai2023sg,zhai20222}, and scene understanding~\cite{busam2015stereo, di2023u,zhai2023commonscenes}. In contrast to traditional instance-level pose estimation methods~\cite{di2021so, labbe2020cosypose}, which rely on specific 3D CAD models for each target object, the category-level approach necessitates greater adaptability to accommodate inherent shape diversity within each category. Effectively addressing intra-class shape variations has thus become a central focus, crucial for real-world applications where objects within a category may exhibit significant differences in shape while sharing the same general category label.

\smallskip
\textbf{Mean Shape \textit{vs.} Semantic Priors.}
One common approach to handle intra-class shape variation involves using explicit mean shapes as prior knowledge \cite{lin2022selfdpdn, tian2020shape, zhang2022rbppose}. These methods typically consist of two functional modules: one for reconstructing the target object by slightly deforming the mean shape and another for regressing the 9D pose based on the reconstructed object~\cite{tian2020shape, lin2022selfdpdn} or enhanced intermediate features~\cite{zhang2022rbppose}.
These methods assume that the mean shape can perfectly encapsulate the structural information of objects within each category, thus achieving reconstruction of the target object with minimal deformation is feasible.
However, this assumption does not hold in reality. 
Objects within the same category, such as chairs, may have fundamental structural differences, leading to the failure of such methods.

Recently, self-supervised learning with large vision models has experienced a significant leap forward, among which DINOv2~\cite{oquab2023dinov2}, due to its exceptional performance in providing semantically consistent patch-wise features, has gained great attention.
In particular, various methods~\cite{zhang2023tale} utilize semantic features from DINOv2 as essential priors to understand the object.
In the field of pose estimation, compared to category-specific mean shapes, DINOv2 demonstrates superior generalization capabilities in object representation across each category, thanks to its large-scale training data and advanced training strategy.
ZSP~\cite{goodwin2022zero} directly leverage DINOv2 features for zero-shot construction of semantic correspondences between objects under different camera viewpoints, and then estimates the pose with RANSAC.
POPE~\cite{fan2023pope} and CNOS~\cite{nguyen2023cnos} harness DINOv2 to refine the object detection, thus implicitly boosting the accuracy of pose estimation.
However, to our knowledge, currently there exists no method that explores how to fuse DINOv2 features with object-specific features to directly enhance the performance of category-level pose estimation.

\begin{figure*}[t]
    \centering
    \includegraphics[width=0.98\textwidth]{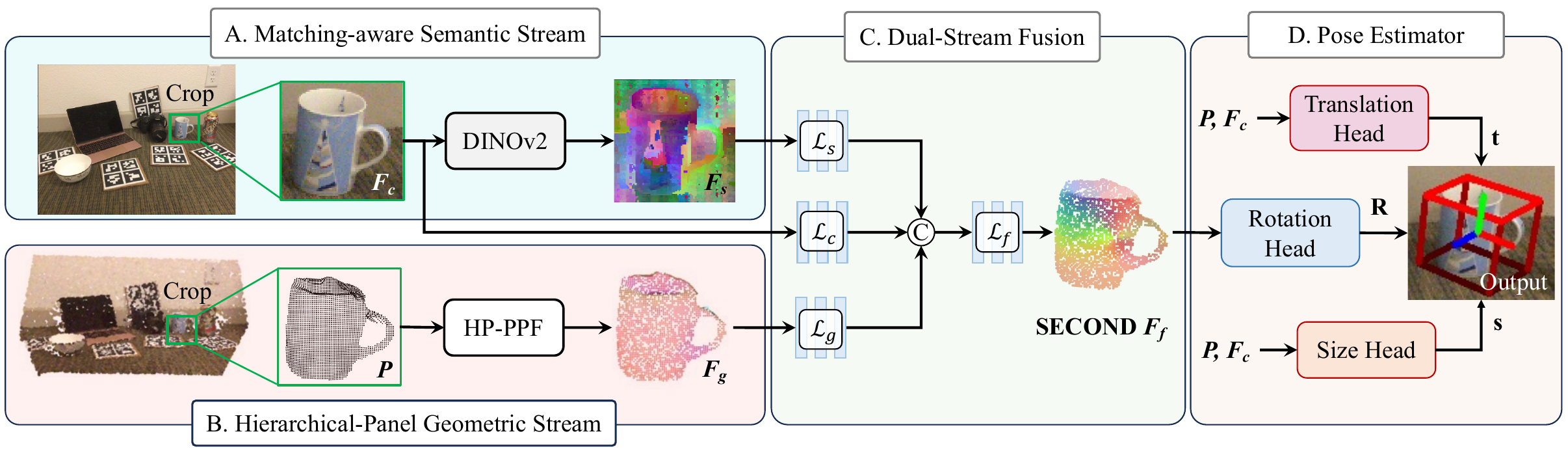}
    \caption{\textbf{Illustration of SecondPose.} Semantic features are extracted using the DINOv2 model (A), and the HP-PPF feature is computed on the point cloud (B). These features, combined with RGB values, are fused into our SECOND feature $F_f$ (C) using stream-specific modules $L_s$, $L_g$, $L_c$, and a shared module $L_f$ for concatenated features. The resulting fused features, in conjunction with the point cloud, are utilized for pose estimation (D).}
    \label{fig:pipe}
\vspace{-0.3cm}
\end{figure*}

In this paper, we present \textbf{SecondPose}, a novel method that fuses \textbf{SE}(3)-\textbf{Con}sistent \textbf{D}ual-stream features to enhance category-level \textbf{Pose} estimation.
Leveraging DINOv2's patch-wise SE(3)-consistent semantic features, we extract two types of SE(3)-invariant geometric features—pair-wise distance and pair-wise angles—to encapsulate object-specific cues.
We hierarchically aggregate geometric features within support regions of increasing radius to encode local-to-global object structure information.
These features are then point-aligned with DINOv2 features to establish a unified object representation that is consistent under SE(3) transformations.
Specifically, given an RGB-D image capturing the target object, we first back-project the depth map to generate the respective point cloud, which is then fed into our \textit{Geometric and Semantic Streams} (Fig.~\ref{fig:pipe}.A-B) to extract the corresponding features for our dual-stream fusion (Fig. \ref{fig:pipe}.C).
The fused features denoted as \textbf{SECOND} are finally fed into an off-the-shelf \textit{pose estimator}~\cite{lin2023vinet} (Fig. \ref{fig:pipe}.D) to regress the 9D pose.

\smallskip
\textbf{SE(3)-Consistent Fusion \textit{vs.} Direct Fusion.}
Alternatively, one could think of directly concatenating DINOv2 features with the back-projected point in a point-wise manner, without extracting SE(3)-invariant geometric features.
However, our instead proposed SE(3)-consistent fusion holds two important advantages over such a straightforward approach.
First, while DINOv2 is trained solely with RGB images, the incorporation of geometric features from the point cloud enriches it with valuable local-to-global 3D structural information. 
This enrichment proves particularly advantageous in handling diverse object shapes within a given category.
Second, our SE(3)-consistent object representation modifies the underlying pose estimation process from \{\textit{point cloud} $\longrightarrow$ \textit{canonical space}\} to \{\textit{point cloud} $\longrightarrow$ \textit{SE(3)-consistent representation} $\longrightarrow$ \textit{canonical space}\}.
In this optimized pipeline, the second stage -- transitioning from our object representation to the human-defined canonical space -- is consistent under SE(3) transformations. (approximately invariant, see Fig.~4)
This consistency significantly simplifies the pose estimation process, as the pose estimator only needs to operate within the second stage. 
Further, this streamlined approach not only enhances the accuracy of pose estimation but also contributes to the efficiency of the overall method.

To summarize, our main contributions are threefold:
\begin{enumerate}
\setlength{\itemsep}{0pt}
\setlength{\parsep}{0pt}
\setlength{\parskip}{1pt}
    \item We present \textbf{SecondPose}, the first method to directly fuse object-specific hierarchical geometric features with semantic DINOv2 features for category-level pose estimation. 
    \item Our SE(3)-consistent dual-stream feature fusion strategy yields a unified object representation that is robust under SE(3) transformations, better suited for down-stream pose estimation. 
    \item Extensive evaluation proves that our SE(3)-consistent fusion strategy significantly boosts pose estimation performance even under severe occlusion and clutter, enabling real-world applications. 
\end{enumerate}
\section{Related Works}

\paragraph{Instance-Level Pose Estimation}
Instance-level pose estimation focuses on determining the 3D rotation and 3D translation of known objects given their 3D CAD models. 
Recent methods can be mainly categorized into three types: direct pose regression~\cite{xiang2018posecnn,kehl2017ssd}, methods that establish 2D-3D correspondences through keypoint detection or pixel-wise 3D coordinate estimation~\cite{wang2021gdr,peng2019pvnet,zakharov2019dpod}, and approaches that learn pose-sensitive embeddings for subsequent pose retrieval~\cite{sundermeyer2018implicit}. 
While most keypoints based approaches rely on the P\textit{n}P algorithm~\cite{peng2019pvnet, zakharov2019dpod, su2022zebrapose} to solve for pose, some methods instead employ neural networks to learn the optimization step~\cite{wang2021gdr}. 
As for RGB-D input, traditional methodologies often rely on hand-crafted features~\cite{drostppf,hinterstoisser2011multimodal}. 
Some more recent approaches~\cite{wang2019densefusion,he2020pvn3d,chen2020g2lnet,he2021ffb6d, wu2022vote} instead extract features independently from RGB images and point clouds, using dedicated CNNs and point cloud networks.
These individual features are then fused for direct pose regression~\cite{wang2019densefusion,chen2020g2lnet} or keypoint detection~\cite{he2020pvn3d, he2021ffb6d, wu2022vote}. 
Despite significant progress, practical applications of these methods remain limited due to their restriction to a few objects and the need for 3D CAD models.

\paragraph{Category-Level Pose Estimation}
In the domain of category-level pose estimation, the objective encompasses predicting the 9DoF pose for any object, regardless if previously seen or novel, from a predefined set of categories. 
This task is inherently more complex due to significant intra-class variations in shape and texture. 
To address these challenges, Wang et al. ~\cite{wang2019normalized} developed the Normalized Object Coordinate Space (NOCS), offering a unified representation framework.
This approach involves mapping the observed point cloud to the NOCS system, followed by pose recovery via the Umeyama algorithm~\cite{umeyama1991least}. 
Alternatively, CASS~\cite{chen2020learning} introduces a learned canonical shape space, while FS-Net~\cite{chen2021fsnet} advocates for a decoupled representation of rotation, focusing on direct pose regression. 
DualPoseNet~\cite{lin2021dualposenet} employs dual networks for both explicit and implicit pose prediction, ensuring consistency for refined pose estimation. 
GPV-Pose~\cite{di2022gpvpose} and OPA-3D~\cite{su2023opa} leverage geometric insights in bounding box projection to augment the learning of pose-sensitive features specific to categories.
HS-Pose~\cite{zheng2023hspose} proposed the HS-layer, a simple network structure that extends 3D graph convolution to extract hybrid scope latent features from point cloud data. 
In contrast, 6-PACK~\cite{wang2021categorylevel} conducts pose tracking by means of semantic keypoints, and CAPTRA~\cite{weng2021captra} combines coordinate prediction with direct regression for enhanced accuracy.
SelfPose~\cite{zaccaria2023self} utilizes optical flow to enhance the pose estimation accuracy.

To address the issue of intra-class shape variations, several works have focused on the incorporation of additional shape priors.
SPD~\cite{tian2020shape} utilizes a PointNet autoencoder to derive a prior point cloud for each category, representing the average shape.
This model is then adapted to fit specific observed instances, assigning the observed point cloud to the reconstructed shape model. 
SGPA~\cite{chen2021sgpa} dynamically adjusts the shape prior based on structural similarities of the observed instances. 
SAR-Net~\cite{lin2022sarnet}, while also employing shape priors, further leverages geometric attributes of objects to enhance performance. 
ACR-Pose~\cite{fan2021acrpose}, instead utilizes a shape prior-guided reconstruction network paired with a discriminator to achieve high-quality canonical representations.

Furthermore, recent research has introduced prior-free methods that demonstrate performance comparable to approaches relying on priors. VI-Net~\cite{lin2023vinet} attains high precision in object pose estimation by separating rotation into viewpoint and in-plane rotations. Additionally, IST-Net~\cite{liu2023istnet} achieves state-of-the-art performance on the REAL275 benchmark by implicitly transforming camera-space features to world-space counterparts without depending on priors.
\section{Method}
The objective of SecondPose is to estimate the 9DoF object pose from a single RGB-D image.
In particular, given an RGB-D image capturing the target object from a set of known categories, our goal is to recover its full 9DoF object pose, including the $\boldsymbol{R} \in S O(3)$ and the $3 \mathrm{D}$ translation $t \in \mathbb{R}^3$ and the $3 \mathrm{D}$ metric size $s \in \mathbb{R}^3$.

\subsection{Overview.}
As illustrated in Fig \ref{fig:pipe}, SecondPose mainly consists of 3 modules to predict object pose from a single RGB-D input, \textit{i.e.} i) the extraction of relevant geometric features $F_g$ and semantic features $F_s$, ii) the dual-stream feature fusion to build our SE(3)-consistent object representation $F_f$, iii) the final pose regression from the extracted representation.

\begin{figure}[t]
  \centering
   \includegraphics[width=\linewidth]{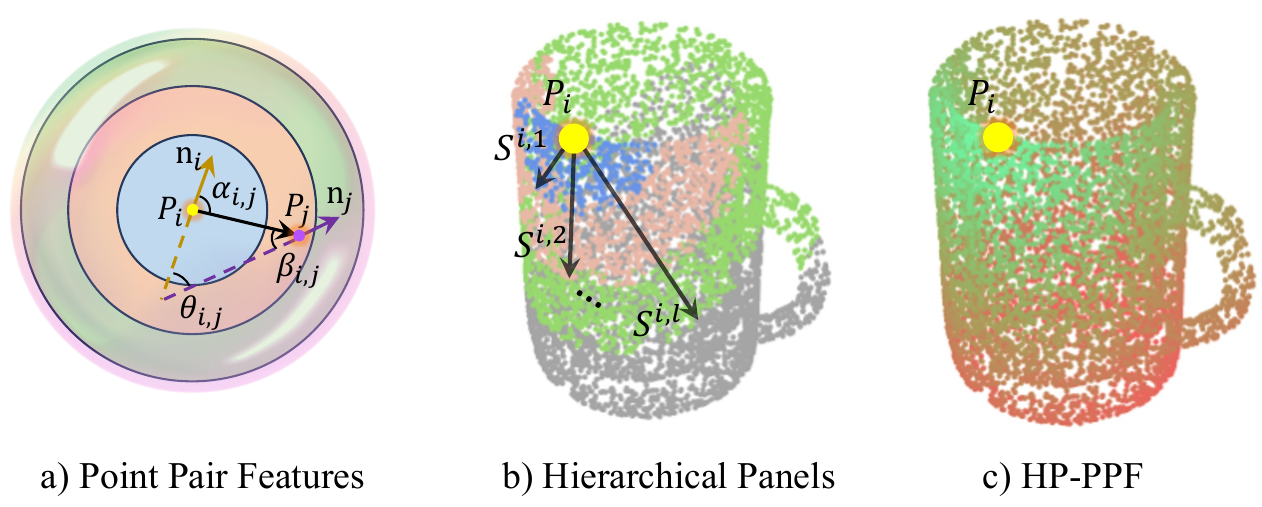}
   \caption{\textbf{Hierarchical panel-based geometric features.} The inner panel contains points that are close to the point of interest, and outer panels contain points far from the point of interest.}
   \label{fig:onecol}
   \vspace{-0.3cm}
\end{figure}

\subsection{Semantic Category Prior From DINOv2}

\paragraph{DINOv2 is an implicit rotation learner} We use DINOv2\cite{oquab2023dinov2} as our image feature extractor. As shown in~\cite{zhang2023tale}, DINOv2 can extract semantic-aware information from RGB images that can be well leveraged to establish zero-shot semantic correspondences, rendering it an excellent method for rich semantic information extraction. 

As for estimating the 3D rotation, such extra semantic-aware information can provide a noticeable boost in performance. 
Exemplary, imagine that the z-axis commonly points to the top side of the object in model space, the y-axis always points to the front side of the object, and the x-axis always points to the left side of the object. Harnessing the semantic information given by DINOv2, the model can more easily identify the top, front, and left sides of the object, thus turning rotation estimation into a much simpler task.
Moreover, DINOv2 features additionally contain global information about the object, including the object category and pose. Such information can thus serve as a good global prior to our method.

\paragraph{Deeper DINOv2 features}
We use the "token" facet from the last (11th.) layer as our extracted semantic feature. Essentially, \cite{zhang2023tale} has demonstrated that the features of deeper layers exhibit optimal semantic matching performance, thus providing improved consistency in terms of semantic correspondence across different objects.  In addition, features from deeper layers also possess more holistic semantic information. A visualization piece is shown in Fig.~\ref{fig:pipe}.A.

\paragraph{Direct pose estimation from DINOv2}
As aforementioned, the ad-hoc fusion of DINOv2 features with the back-projected points exhibits several downsides. First, DINOv2 extracts information only from RGB images; hence, the contained geometric information is limited. Second, as we make use of deeper-layer features from DINOv2 for a more holistic representation, the local detailed information is blurred to some extent. To complement DINOv2 features in these aspects, we thus need to combine them with geometric features containing local information for better descriptive power.

\subsection{Hierarchical Geometric Features}
The stream pipeline is shown in Fig.~\ref{fig:pipe}.B. Our geometric embedding in this stream is based on the calculation of pair-wise SE(3)-equivariant Point Pair Features (PPFs)~\cite{drostppf}. We construct our SE(3)-invariant coordinate representation by aggregating the PPFs between the point of interest and neighborhood points in the multiple panels centered on it. We hierarchically concatenate the corresponding SE(3)-invariant coordinate representations in each panel to enrich the representation power of our geometric features HP-PPF. Fig.~\ref{fig:onecol}.c provides a visualization of HP-PPF.

\paragraph{Point Pair Features PPFs} A comprehensive example is shown in Fig.~\ref{fig:onecol}.a. Given an object point cloud denoted as $\boldsymbol{P}$, we consider each pair of points $(p_i, p_j)$ where $p_i, p_j \in \boldsymbol{P}$. Associated with each point, local normal vectors $\boldsymbol{n_i}$ and $\boldsymbol{n_j}$ are computed at each $p_i$ and $p_j$, respectively. The final pairwise feature between $p_i$ and $p_j$ is defined as

\begin{equation}
f_{i,j} = [d_{i,j}, \alpha_{i,j}, \beta_{i,j}, \theta_{i,j}],
\label{eq:ppf}
\end{equation}
where $d_{i,j} = \lVert p_j - p_i \rVert$ describes the Euclidean distance between points $p_i$ and $p_j$. $\alpha_{i,j} = \angle (\boldsymbol{n_i}, p_j-p_i)$ represents the angular deviation between the normal vector $\boldsymbol{n_i}$ at point $p_i$ and the vector extending from $p_i$ to $p_j$. $\beta_{i,j} = \angle (\boldsymbol{n_j}, p_j-p_i)$ denotes the angle subtended by the normal vector $\boldsymbol{n_j}$ at point $p_j$ with the aforementioned vector from $p_i$ to $p_j$. $\theta_{i,j} = \angle (\boldsymbol{n_j}, \boldsymbol{n_i})$ denotes the angular disparity between the normal vectors $\boldsymbol{n_j}$ and $\boldsymbol{n_i}$ at points $p_j$ and $p_i$, respectively.
Notice that thanks to its locality, this descriptor is invariant under $SE(3)$.

\paragraph{Geometric Feature Panel}
Based on PPFs, we propose panel-based PPFs to construct our geometric representation, which increases the perception field while maintaining the merit of the locality. For each point $p_i$ in the point cloud $\boldsymbol{P}$, there is a support panel $\mathcal{S}^i \subseteq \boldsymbol{P}$ whose cardinality $s_i = |\mathcal{S}^i|$. For all points $p_j \in \mathcal{S}^i$, we calculate the PPF $f_{i, j}$ between $p_i, p_j$ and the local coordinate representation $f^i_l$ of $p_i$ is then obtain as average the average according to
\begin{equation}
f^i_l = \frac{1}{s_i}(\sum_j{d_{i,j}}, \sum_j{\alpha_{i,j}}, \sum_j{\beta_{i,j}},\sum_j{\theta_{i,j}}).
\end{equation}

\paragraph{From Single to Hierarchical Panels}
Even though the mean aggregation in the panel can take the neighboring points into account, the inherent local representation limits its representational power, as the features brought by normals $\boldsymbol{n_i},\boldsymbol{n_j}$ are noisy when constraining the perception field. Inspired by CNNs, which extract hierarchical features from local to global, we  hierarchically sample multiple panels from local to global, as shown in Fig.~\ref{fig:onecol}.b.
Specifically, for a point set $\boldsymbol{P}$ with cardinality $|\boldsymbol{P}|$, for integers $(k_0, k_1, k_2,..., k_l)$ satisfying $ 0 = k_0<k_1<k_2<...<k_l=|\boldsymbol{P}|-1$,  for each point $p_i\in \boldsymbol{P}$ we first rank its distance to any other points in $\boldsymbol{P}$ from smallest to largest:
\begin{equation}
    r_{i,j} = sort(d_{i,j})
\end{equation}
and construct support panels:
\begin{equation}
  \mathcal{S}^{i,m}  = \{p_j \in \boldsymbol{P}| k_{m-1}< r_{i,j}\leq  k_m \}, 1 \leq m \leq l,
\end{equation}
with $l$ being the number of employed panels. We then calculate the corresponding pose-invariant coordinate representations $f^{i,m}$ for each panel $\mathcal{S}^{i,m}$ and concatenate them to get the point-wise geometric features with
\begin{equation}
    f_{g}^i = f^{i,1}_l\oplus f^{i,2}_l\oplus  ...\oplus f^{i,l}_l .
\end{equation}
Thereby, for smaller $k$, the support panel is composed of points that are closer to the point of interest, whereas for larger $k$, the support panel consists of points that are farther from the point of interest. By concatenating features calculated by panels of different scales, we can harness geometric features in a way that balances details of local geometric landscapes and global instance-wise shape information. We experimentally show in Sec.~\ref{exper} that our design performs better than the usual single-panel descriptor.

\subsection{SE(3)-Consistent Feature Fusion}

\paragraph{Fusion Strategy} We fuse the DINOv2 features, the geometric feature and RGB values, as shown in Fig.~\ref{fig:pipe}.C. In particular, we use VI-Net~\cite{lin2023vinet} as an example of the pose estimator, first projecting each feature to each feature stream $\mathcal{F}$ and 3D point cloud $P = \{p_i\}$ to a spherical feature map $F$. To this end, we divide the sphere uniformly into $W \times H$ along the azimuth and elevation axes, following VI-Net~\cite{lin2023vinet}. We assign the feature of the point with the largest distance to each bin.
When there is no point in the region, we set 0 in the bin.
For each feature map $F_i \in \{F_g, F_s, F_c \}$ representing the geometric feature, the DINOV2 feature, and the respective RGB value, we employ a separate ResNet model $\mathcal{L}_i$ as feature extractor. The outputs of these individual feature extractors are then concatenated to form the input to another ResNet for feature fusion, obtaining $F_{f}$ also denoted as \textbf{SECOND},
\begin{equation}
F_{f} = \mathcal{L}_f\left(\mathcal{L}_g(F_g) \oplus \mathcal{L}_s(F_s) \oplus \mathcal{L}_c(F_c)\right).
\end{equation}
\paragraph{Advantages of SE(3)-Consistent Fusion} The Design of a SE(3)-consistent fusion is an integral part of the improved quality of our method.
As for the 3D rotation, we are learning a mapping from the space of point clouds and its features $(P,F) \in \mathbb{R}^{n \times 3} \times \mathbb{R}^{n \times C} $ to space of 3D rotations $\boldsymbol{R} \in S O(3)$
\begin{equation}
    \Phi: \mathbb{R}^{n \times 3} \times \mathbb{R}^{n \times C} \mapsto S O(3).
\end{equation}
This mapping $\Phi$ should ensure rotation-equivariance, meaning that
\begin{equation}
    \Phi(R_xP, \psi_{R_x} (F)) = R_x\Phi(P,F), \forall R_x \in  S O(3),
    \label{re}
\end{equation}
where $\psi_{R_x}$ is the transformation applied to the feature when rotating the point cloud by $R_x$.
This rotation-equivariance relation is essential for the learned model to generalize well on unseen data. Without such equivariance embedded in the model structure, these relation needs to be learned through large amounts of data, which is limited by the scale of the data. Our design of SE(3)-consistent features are approximately rotation-invariant, hence
\begin{equation}
    \psi_{R_x}(F) \approx F , \forall R_x \in  S O(3),
\end{equation}
eliminating the effect of $\psi_{R_x}$ in Eq.~\eqref{re}, and thus making learning of the rotation-equivariance relationship easier.

\subsection{SecondPose Training and Inference}
Following \cite{lin2023vinet}, we leverage a lightweight PointNet++~\cite{qi2017pointnet} as the translation and size estimation heads. Given an RGB-D 
image, we first segment the object of interest using Mask-RCNN \cite{he2017mask}, similar to \cite{di2022gpvpose, lin2023vinet}. We then randomly select $N$ points from the back-projected 3D point clouds $\boldsymbol{P} \in \mathbb{R}^{n \times 3}$ with RGB features $F_c$ and use them to estimate the translation and size, as shown in Fig.~\ref{fig:pipe}.D.

The core of our method is thus developed to focus on the more challenging task of 3D rotation estimation.
We essentially train a separate translation-size network and rotation network. For the translation-size network, we adopt the L1 loss for both size and translation with
\begin{equation}
 L_{ts} = \lambda_t |t_{pred} - t_{gt}| +  \lambda_s |s_{pred} - s_{gt}|.
 \end{equation}
 For the 3D rotation, we instead directly predict the 9D rotation matrix, which we optimize via the L1-loss according to
 \begin{equation}
 L_{R} = |R_{pred} - R_{gt}|. 
 \end{equation}
During training, the ground truth translation and size are used to center and normalize the point cloud before rotation estimation, while during inference the predicted size and translation are instead utilized for normalization.

\begin{table*}[!htpb]
\centering
\resizebox{0.8\linewidth}{!}{
\begin{tabular}{lcccccc}
\toprule
Method  & Mean & \multicolumn{5}{c}{ REAL275 } \\
\cline { 3 - 7 } Name & Shape Priors & IoU $_{75} *$ & $5^{\circ}\ 2 \mathrm{~cm}$ & $5^{\circ}\ 5 \mathrm{~cm}$ & $10^{\circ}\ 2 \mathrm{~cm}$ & $10^{\circ}\ 5 \mathrm{~cm}$ \\
\midrule
SPD \cite{tian2020shape} & $\checkmark$ & 27.0 & 19.3 & 21.4 & 43.2 & 54.1 \\
CR-Net \cite{wang2021categorylevel} & $\checkmark$ & 33.2 & 27.8 & 34.3 & 47.2 & 60.8 \\
CenterSnap-R \cite{irshad2022centersnap} &$\checkmark$  & - & - & 29.1 & - & 64.3 \\
ACR-Pose \cite{fan2021acrpose} & $\checkmark$ & - & 31.6 & 36.9 & 54.8 & 65.9 \\
SAR-Net \cite{lin2022sarnet} &$\checkmark$  & - & 31.6 & 42.3 & 50.3 & 68.3 \\
SSP-Pose \cite{zhang2022ssppose} & $\checkmark$ & - & 34.7 & 44.6 & - & 77.8 \\
SGPA \cite{chen2021sgpa} & $\checkmark$ & 37.1 & 35.9 & 39.6 & 61.3 & 70.7 \\
RBP-Pose \cite{zhang2022rbppose} &$\checkmark$  & - & 38.2 & 48.1 & 63.1 & 79.2 \\
SPD + CATRE \cite{liu2022catre} & $\checkmark$ & 43.6 & 45.8 & 54.4 & 61.4 & 73.1 \\
DPDN \cite{lin2022categorylevel} &$\checkmark$  & - & 46.0 & 50.7 & $70.4$ & 78.4 \\
\midrule
FS-Net \cite{chen2021fsnet} & $\times$ & - & - & 28.2 & - & 60.8 \\
DualPoseNet \cite{lin2021dualposenet} & $\times$ & 30.8 & 29.3 & 35.9 & 50.0 & 66.8 \\
GPV-Pose \cite{di2022gpvpose} & $\times$ & - & 32.0 & 42.9 & - & 73.3 \\
SS-ConvNet \cite{Lin2021SparseSC} & $\times$ & - & 36.6 & 43.4 & 52.6 & 63.5 \\
HS-Pose \cite{zheng2023hspose}  & $\times$ & - & 46.5 & 55.2 & 68.6 & $\underline{82.7}$ \\
IST-Net \cite{liu2023istnet}  & $\times$ & - & 47.5 & 53.4 & $\underline{72.1}$ & 80.5 \\
VI-Net \cite{lin2023vinet}  & $\times$ & $\underline{48.3}$ & $\underline{50.0}$ & $\underline{57.6}$ & 70.8 & 82.1 \\
\midrule
\textbf{SecondPose (Ours)} & Semantic Priors & $\textbf{49.7}$ & $\textbf{56.2}$ & $\textbf{63.6}$ & $\textbf{74.7}$ & $\textbf{86.0}$ \\
\bottomrule
\end{tabular}
}
\caption{Quantitative comparisons of different methods for category-level 6D object pose estimation on REAL275~\cite{wang2019normalized}. ‘*’ denotes the CATRE\cite{liu2022catre} IoU metrics. The best results are in bold, and the second best results are underlined.}
\label{tab:mainresults_real275}
\vspace{-0.5cm}
\end{table*}

\section{Experiment}
\label{exper}
\subsection{Experimental Setup.}
\paragraph{Datasets} We conduct our experiments on the common 9D pose estimation benchmarks  NOCS-REAL275~\cite{wang2019normalized}, NOCS-CAMERA25~\cite{wang2019normalized} as well as HouseCat6D\cite{jung2022housecat6d} datasets. 
NOCS-REAL275 is a real-world dataset with 13 scenes containing objects from 6 different categories; 4,300 images of 7 scenes are used as a training set, while the other 2,750 images of 6 scenes form the test set.
NOCS-CAMERA25 is a synthetic dataset containing 300k images with objects from the same categories as NOCS-REAL275.
HouseCat6D is a comprehensive multi-modal real-world dataset, featuring 194 high-fidelity 3D models of household items of 10 categories. 
The collection encompasses transparent and reflective objects situated in 41 scenes, presenting a wide range of viewpoints, challenging occlusions, and devoid of markers. 

\paragraph{Evaluation Metrics} 
As for the NOCS-REAL275 dataset, we report the mean Average Precision (mAP) of $5^{\circ} 2 \mathrm{cm}, 5^{\circ} 5 \mathrm{cm}, 10^{\circ} 2 \mathrm{cm}, 10^{\circ} 5 \mathrm{cm}$ metrics. $n^{\circ} m \mathrm{cm}$ denotes the percentage of prediction with rotation prediction error within n degrees and translation prediction error within m centimeters.
We also report mAP of 3D Intersection over Union (IoU) at the threshold of $75 \%$.
For the HouseCat6D dataset, we again report the mAP of 3D IoU under thresholds of $25 \%$ and $50 \%$. 
\begin{figure*}[t]
  \centering
   \includegraphics[width=0.98\linewidth]{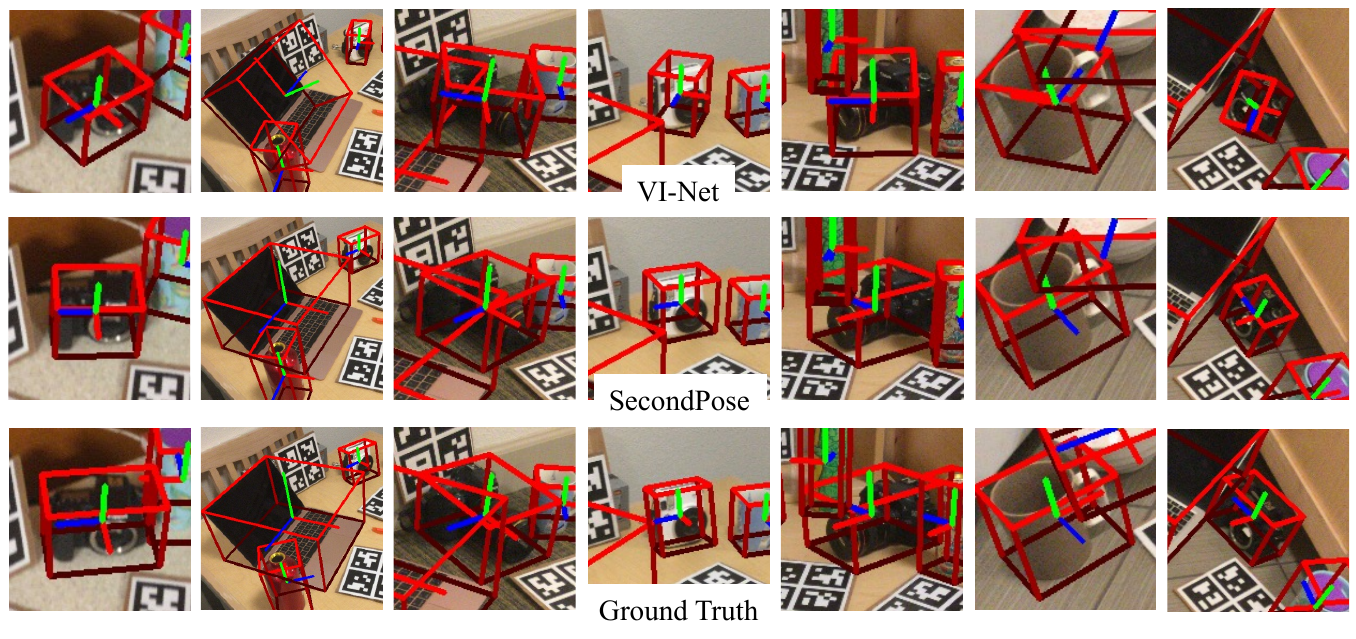}

   \caption{Qualitative comparison on REAL275~\cite{wang2019normalized}. We compare our prediction with ground truth and the prediction of our baseline, VI-Net\cite{lin2023vinet}. Our approach achieves significantly higher precision in rotation estimation. }
   \label{fig:qualitative}  
   \vspace{-0.2cm}
\end{figure*}

\paragraph{Efficiency} Our method achieves an inference speed of 9 FPS. Excluding the running time of DINOv2, our inference speed increases to 10 FPS.

\paragraph{Implementation Details}
We use MaskRCNN\cite{he2017mask} to segment the objects of interest from the input image. 
We then combine point-wise radial distances, RGB values, and semantic-aware features from DINOv2 together with our proposed local-to-global SE(3)-invariant geometric features as input for further processing.
Next, for the RGB values and the point-wise radial distances, we sample 2048 points from the point cloud. 
For DINOv2 features, we first crop the image by the bounding box around the object of interest and then resize the image to a resolution of $210 \times 210$. Finally, for our geometric features, we sample 300 points from previously sampled 2048 points and estimate point-wise normal vectors using the 10 nearest neighbors. 
To train our model on the NOCS dataset, we use a mixture of 25\% real-world images from the training set of REAL 275 and 75\% synthetic images from the CAMERA25 training set, similar to~\cite{wang2019normalized}. 
For all experiments, we train our models with batch size 48 on a single NVIDIA 3090 GPU to the 40th. epoch.
\begin{figure}[t]
  \centering
   \includegraphics[width=0.98\linewidth]{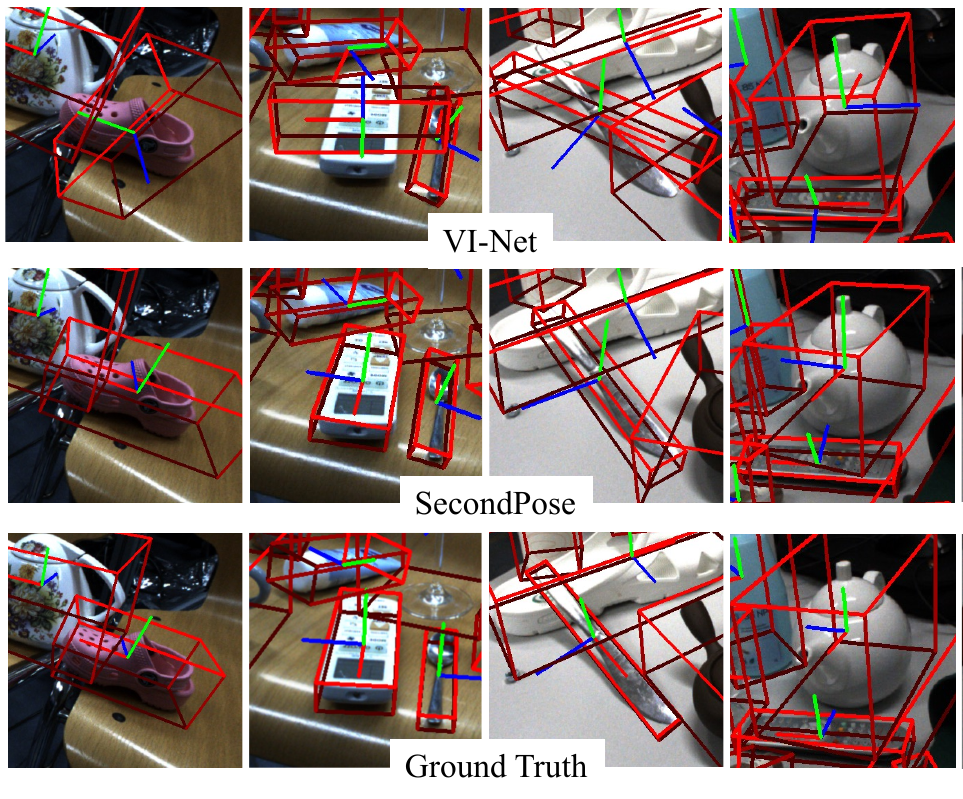}
   \caption{Qualitative comparison on HouseCat6D~\cite{jung2023housecat6d}. We compare our prediction with ground truth and the prediction of our baseline, VI-Net\cite{lin2023vinet}.}
   \label{fig:qualitative_housecat}
   \vspace{-0.5cm}
\end{figure}
\subsection{Comparison with State-of-the-Art Methods} 
In Tab.~\ref{tab:mainresults_real275}, we compare SecondPose with the state-of-the-art on NOCS-REAL275 dataset. 
As can be easily observed, our method outperforms all state-of-the-art approaches, including the recent
VI-Net~\cite{lin2023vinet}, by a large margin on all metrics.
More specifically, our method respectively exceeds VI-Net for $5^{\circ} 2 \mathrm{~cm}$ and $10^{\circ} 2 \mathrm{~cm}$ by 6.2\% and 3.9\%, demonstrating the effectiveness of our SE(3)-consistent feature fusion design.
When comparing with DPDN~\cite{lin2022selfdpdn}, the best method using mean shape prior, our improvements in $5^{\circ} 2 \mathrm{~cm}$ and $5^{\circ} 5 \mathrm{~cm}$ metrics amount to 10.2\% and 12.8\%.
We show qualitative results in Figure~\ref{fig:qualitative}.
It can be observed that SecondPose is more robust when handling objects with large intra-class variations, such as \textit{camera}.
In Tab.~\ref{tab:mainresults_housecat}, we evaluate our method on the HouseCat6D dataset.
Our method can again exceed current state-of-the-art methods by a large margin. 
As for the $\text{IoU}_{50}$ metric, our method outperforms the second-best method VI-Net by 9.7\% on average.
Additional qualitative results can be found in Fig.

\begin{table*}[!htpb]
  \centering
  \resizebox{0.95\linewidth}{!}{
  \begin{tabular}{l|c|cccccccccc}
\toprule
\text{ Approach } & $\text{IoU}_{25}$ / $\text{IoU}_{50}$ & \text { Bottle } & \text { Box } & \text { Can } & \text { Cup } & \text { Remote } & \text { Teapot } & \text { Cutlery } & \text { Glass } & \text { Tube } & \text { Shoe } \\
\midrule
\text { NOCS \cite{wang2019normalized} } & 50.0 / 21.2 & 41.9 / 5.0 & 43.3 / 6.5 & 81.9 / 62.4 & 68.8 / 2.0 & 81.8 / 59.8 & 24.3 / 0.1 & 14.7 / 6.0 & 95.4 / 49.6 & 21.0 / 4.6 & 26.4 / 16.5 \\
\text { FS-Net \cite{chen2021fsnet} } & 74.9 / 48.0 & 65.3 / 45.0 & 31.7 / 1.2 & 98.3 / 73.8 & 96.4 / 68.1 & 65.6 / 46.8 & 69.9 / 59.8 & 71.0 / 51.6 & 99.4 / 32.4 & 79.7 / 46.0 & 71.4 / 55.4 \\
\text { GPV-Pose \cite{di2022gpvpose} } & 74.9 / 50.7 & 66.8 / 45.6 & 31.4 / 1.1 & 98.6 / 75.2 & 96.7 / 69.0 & 65.7 / 46.9 & 75.4 / 61.6 & 70.9 / 52.0 & 99.6 / 62.7 & 76.9 / 42.4 & 67.4 / 50.2 \\
\text { VI-Net \cite{lin2023vinet} } & 80.7 / 56.4 & 90.6 / 79.6 & 44.8 / 12.7 & $\textbf{99.0}$ / 67.0 & 96.7 / 72.1 & \textbf{54.9} / 17.1 & 52.6 / 47.3 & 89.2 / \textbf{76.4} & 99.1 / \textbf{93.7} & \textbf{94.9} / \textbf{36.0} & 85.2 / 62.4 \\
\midrule
\textbf { SecondPose (Ours) } & $\textbf{83.7}$ / $\textbf{66.1}$ & $\textbf{94.5}$ / $\textbf{79.8}$ &$\textbf{54.5}$  / $\textbf{23.7}$ & $98.5$ / $\textbf{93.2}$ & $\textbf{99.8}$ / $\textbf{82.9}$ & 53.6 / $\textbf{35.4}$ & $\textbf{81.0}$ / $\textbf{71.0}$ & $\textbf{93.5}$ / 74.4 & $\textbf{99.3}$ /92.5  & 75.6 / 35.6 &$\textbf{86.9}$  / $\textbf{73.0}$ \\
\bottomrule
 \end{tabular}
 }
  \caption{Overall and class-wise evaluation of 3D IoU(at 25\%, 50\%) on the dataset HouseCat6D~\cite{jung2023housecat6d}. The best results are in bold.}
  \label{tab:mainresults_housecat}
  \vspace{-0.3cm}
\end{table*}

\begin{table*}[t]
  \centering
\small
\begin{tabular}{ll|c|cccc}
\toprule
\text { Row } & \text { Method }  & $\mathrm{IoU}_{75}*$ & $5^{\circ}\ 2\mathrm{~cm}$ & $5^{\circ}\ 5\mathrm{~cm}$ & $10^{\circ}\ 2\mathrm{~cm}$ & $10^{\circ}\ 5\mathrm{~cm}$ \\
\midrule
\text { A0 } & \text { SecondPose  (baseline) } & 49.7 & 56.2 & 63.6 & 74.7 & 86.0 \\
\midrule
\text { B0 } & \text { w/o semantic }  & 48.0 & 51.1 & 58.9 & 71.6 & 82.4 \\
\text { B1 } & \text { w/o geometric }  & 49.5 & 55.1 & 62.3 & 73.7 & 84.8 \\
\text { B2 } & \text { w/o semantic+geometric } & 48.5 & 49.9 & 57.4 & 70.4 & 80.8 \\
\midrule
\text { C0 } & \text { w/o d in Eq. \ref{eq:ppf} }  & 49.1 & 55.1 & 63.1 & 73.7 & 85.0 \\
\text { C1 } & \text { w/o $\alpha$ in Eq. \ref{eq:ppf} }  & 49.3 & 54.7 & 62.8 & 73.1 & 84.7 \\
\text { C2 } & \text { w/o $\beta$ in Eq. \ref{eq:ppf} }  & 49.6 & 54.8 & 62.7 & 74.6 & 86.7 \\
\text { C3 } & \text { w/o $\theta$ in Eq. \ref{eq:ppf} }  & 49.5 & 55.1 & 63.1 &  74.2 & 85.6 \\
\midrule
\text { D0 } & \text { KNN Panel (10 nearest neighbors) }  & 49.4 & 55.4 & 63.1 &  73.7 & 85.5\\
\midrule
\text { E0 } & \text { random rotation }$5^{\circ}$  & 49.7 & 56.1 & 63.4 &  74.6 & 85.9\\
\text { E1 } & \text { random rotation }$10^{\circ}$   & 49.4 & 55.8 & 63.5 &  74.4 & 85.8\\
\text { E2 } & \text { random rotation }$15^{\circ}$   & 48.5 & 55.4 & 63.0 &  73.9 & 85.4\\
\text { E3 } & \text { random rotation }$20^{\circ}$  & 47.9 & 54.5 & 62.4 &  73.2 & 85.1\\
\midrule
\text { F0 } & \text { manual occlusion n = 16}  & 49.7 & 56.0 & 63.6 &  74.8 & 86.2\\
\text { F1 } & \text  { manual occlusion n = 8}    & 49.5 &  55.7 & 63.2 &  74.3 &  85.6\\
\text { F2 } & \text  { manual occlusion n = 4}   &  46.7 & 52.5 & 60.9 &  71.5 & 84.6\\
\midrule
\text { G0 } & \text { random perturbation s = 0.002}  & 49.7 & 56.1 & 63.6 &  74.6 & 85.8\\
\text { G1 } & \text { random perturbation s = 0.005}  & 49.6 & 55.8 & 63.4 &  74.4 & 86.0\\
\text { G2 } & \text { random perturbation s = 0.01}  & 45.9 & 53.7 & 62.6 &  73.4 & 86.1\\

\bottomrule
\end{tabular}
  \caption{Ablation Study on REAL275~\cite{wang2019normalized}. ‘*’ denotes the CATRE\cite{liu2022catre} IoU metrics.}
  \label{tab:ablation_train}
  \vspace{-0.5cm}
\end{table*}

\subsection{Limitations}
Our method's efficacy is restricted by the constraints of DINOv2 due to our utilization of its features. When DINOv2 is unable to provide meaningful semantic information for specific images, our approach is unable to surpass this limitation. 

\subsection{Ablation Studies}
To confirm the efficacy of our design choices, we conduct several ablation studies on the NOCS-REAL275\cite{wang2019normalized} dataset. 

\textbf{[AS-1] Efficacy of employing semantic and geometric features.}
To show the effectiveness of our semantic-geometric-feature-fusion, we train the proposed model in 3 different variations: i) without semantic feature, ii) without geometric feature, and iii) without both semantic and geometric features. 
The results are presented in Tab. \ref{tab:ablation_train} (B0) - (B2). 
When considering the strict $5^{\circ} 2 \mathrm{cm}$ metric, it turns out that removing semantic features, geometric features or both always leads to a large decrease in performance. In particular, the performance respectively drops by 5.1\%,  1.1\% and 6.3\%.

\textbf{[AS-2] Efficacy of individual geometric feature.} 
We further run ablations on the four geometric features, $d$, $\alpha$, $\beta$, $\theta$. 
The corresponding results are presented again in Table \ref{tab:ablation_train} under (C0) - (C3). 
As can be observed, removing any component from the geometric feature leads to a strict drop in performance. Exemplary, for the $5^{\circ} 2 \mathrm{cm}$ metric the performance drops by at least 1.1\%. 
To summarize, each geometric feature contributes to the expressiveness of the geometric representation.

\textbf{[AS-3] Efficacy of hierarchical panel construction.} 
As shown in Tab.~\ref{tab:ablation_train} (D0), when the hierarchical panel is substituted by KNN with 10 nearest neighbors, the $5^{\circ} 2 \mathrm{cm}$ metric undergoes a decrease by 0.8\%. 
This demonstrates the importance of our hierarchical panel construction, as it better captures finer-grained local and global information.

\textbf{[AS-4] Robustness under random rotation.}
To show the robustness of our method under random rotation applied on point cloud, we perform experiments on test images when randomly rotating the entire point cloud by rotation angle $A~[0^{\circ}, n^{\circ}]$, n = 5, 10, 15, 20, see Table \ref{tab:ablation_train} (E0) - (E3). 
The results show that our method performs well under these circumstances.

\textbf{[AS-5] Robustness under manual occlusions} 
We also perform an additional experiment to show the robustness of our method under various levels of occlusions. 
We manually mask out the object with different scale of rectangular masks, whose length and width is set to 1/n of the length and width of the original object bounding box. 
We further run tests with n = 16, 8, 4 in Tab. \ref{tab:ablation_train} (F0) - (F2).
When undergoing only mild occlusion, \emph{i.e.} $n=16$, the performance is almost identical to the original result.
Moreover, even when dealing with very large occlusions of 1/4 of the size of the object, the performance is still fairly strong with only a small decrease of 3\% for $\text{IoU}_{75}$.

\textbf{[AS-6] Robustness Under Perturbation on Point Cloud.} 
Next, we also evaluate the robustness of our method under random perturbations applied to the point clouds. 
To this end, we add random noise sampled from a uniform distribution ranging from $-0.5sr$ to $0,5sr$, where $s$ is the scale factor and r is the average distance of the point cloud to the object center. 
We test our model with s = 0.002, 0.005,0.01 in Table \ref{tab:ablation_train} G0-G2. We observe again that with mild perturbation of s = 0.002, the performance is almost identical to the original result, while with relatively large perturbation of s = 0.01, the performance is still fairly strong with only a small decrease of 3.8\% for $\text{IoU}_{75}$.

\section{Conclusion}
In this paper, we propose SecondPose designing SE(3)-consistent fusion of semantic and geometric features for pose estimation. The two feature streams are proven to complement each other and jointly contribute to improving our method. To confirm the efficacy of our method, we apply our method on the challenging real-world category-level 6D object pose estimation datasets REAL275 and HouseCat6D and exceed the current SOTA by a large margin.


{\small
\bibliographystyle{ieee_fullname}
\bibliography{egbib}
}

\newpage
\appendix
\section*{Supplementary Material}

\section{Implementation Details}
Our network is implemented on Pytorch 1.13. The backbone is based on VI-Net~\cite{lin2023vinet}.
To obtain DINOv2 features, we initially crop the object by its bounding boxes from the original image and subsequently resize it to a resolution of $210 \times 210$. The DINOv2 model version employed is 'dinov2\_vits14', with a set stride of 14. Consequently, the resolution of the output DINOv2 feature is $15 \times 15$. We randomly select 100 points from the feature map as our sampled points with DINOv2 features.

For extracting geometric features, we initially randomly sample 300 points from the entire point cloud. These points serve as the basis for estimating point-wise normal vectors. To create the hierarchical panels, we then choose range parameters $(k_0, k_1, 
k_2,k_3, k_4, k_5, k_6) = (0, 10, 20, 40, 80, 160, 299)$.

See Tab \ref{tab:parameter_count} for an overview of the number of trainable parameters and frozen parameters of our method and VI-Net.

\begin{table}[h!]
\vspace{-0.3cm}
  \centering
  \scalebox{0.77}{
  \begin{tabular}{@{}lcc@{}}
    \toprule
    Number of Parameters & Trainable & Frozen  \\
    \midrule
    VI-Net & 27,311,368 & 0  \\
   Ours & 33,639,561 & 22,056,576  \\ 
    \bottomrule
  \end{tabular}}
  \caption{Parameter Count}
  \label{tab:parameter_count}
\end{table}
\vspace{-0.5cm}

\section{Further Explanations of the Pipeline}
\textbf{Invariance \textit{vs} Equivariance.} Following VI-Net\cite{lin2023vinet}, our backbone ensures that, when the input point-wise features are approximately SE(3)-invariant, the output feature map is approximately SE(3)-equivariant. We used the term ``SE(3)-invariant" to emphasize that our input features are invariant. The process of feature fusion is illustrated in Figure 1 below. The RGB values $F_c$, the DINOv2 features $F_s$, and the HP-PPF features $F_g$ are our invariant input features.\\
1) All input features are approximately invariant: the geometric features and RGB values are inherently invariant. The DINOv2 features are approximately invariant due to their training on a large-scale dataset, ensuring consistent semantic representation. This consistency in semantic meaning, regardless of rotation/translation, implies SE(3) consistency, thus leading to approximate SE(3) invariance.\\
2) The output is equivariant: similar to VI-Net, our backbone transforms point-wise features with the point cloud's 3D coordinates into a 2D feature map. These maps are then processed by ResNets that approximately maintain SE(3) equivariance (see section 3.4 and \cite{lin2023vinet} for details). Consequently, when the input features are approximately SE(3) invariant, the output 2D feature map is approximately SE(3) equivariant. 

\begin{figure}[t]
    \centering
    \includegraphics[width=\linewidth]{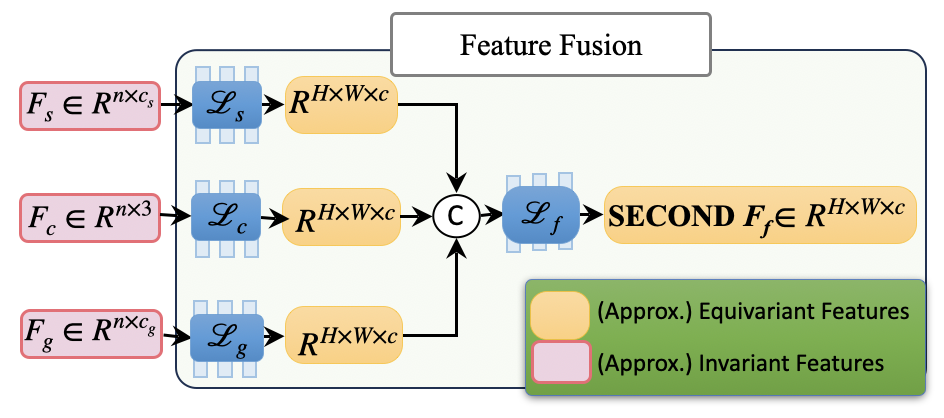}
    \caption{\textbf{Feature Fusion} We illustrate the fusion process with annotated approximately equivalent and approximately invariant features. }
    \label{fig:fusion}
\end{figure}

\begin{figure}[h!]
\vspace{-0.3cm}
    \centering
    \includegraphics[width=0.45\textwidth]{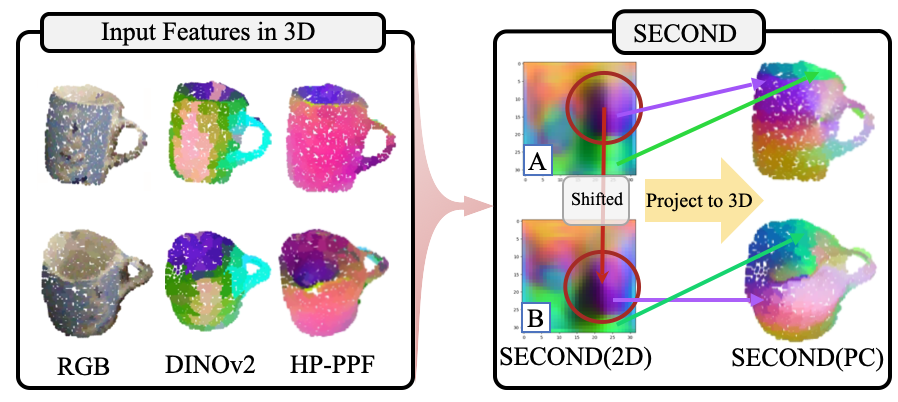}
    \caption{\textbf{Feature Maps} We fuse features of RGB, DINOv2 and HP-PPF into a 2D feature map that is approximately SE(3)-equivariant.}
    
    \label{fig:features}
\vspace{-0.5cm}
\end{figure}

\textbf{Visualization of Feature Maps.} In Fig \ref{fig:features}, we visualize the features of the same object in two frames, each with a different pose.

``RGB", ``DINOv2", and ``HP-PPF" depict our input features, which are roughly SE(3) invariant. ``Second(2D)" represents our post-fusion feature map, utilized for pose estimation and approximately SE(3) equivariant; we observe slight shifts in the feature map pattern upon rotation and translation. For visualization, we also present ``Second(PC)", a point-wise feature obtained by projecting ``Second(2D)" back onto the point cloud, using the pixel-point correspondence, and it's also approximately invariant. 

\textbf{Shape of All Feature Maps and Other Intermediate Representation.} Fig~\ref{fig:fusion} illustrates our input features as:  $F_c \in R^{n\times 3}$, $F_g \in R^{n\times c_g}$, and $F_s \in R^{n\times c_s}$, while after modules $\mathscr{L}_c$, $\mathscr{L}_g$, $\mathscr{L}_s$, our features have shape $R^{H\times W \times c}$, and after fusion module $\mathscr{L}_f$ the feature is of shape $R^{H\times W \times c}$.

\section{More Experimental Results on HouseCat6D}
We report more metrics on HouseCat6D~\cite{jung2022housecat6d} in Table \ref{tab:mainresults_housecat2}. We note that our approach outperforms other methods by a significant margin across all metrics. Especially on the restricted metrics $\text{IoU}_{75}$ and $5^{\circ}\ 2 \mathrm{~cm}$, SecondPose outperforms VI-Net by 22.1\% and 31.0\% respectively.

\begin{table*}[!htpb]
\centering
\begin{tabular}{lccccc}
\toprule
\multirow{2}{*}{Method}   & \multicolumn{5}{c}{ HouseCat6D } \\
  & IoU $_{75}$ & $5^{\circ}\ 2 \mathrm{~cm}$ & $5^{\circ}\ 5 \mathrm{~cm}$ & $10^{\circ}\ 2 \mathrm{~cm}$ & $10^{\circ}\ 5 \mathrm{~cm}$ \\
\midrule
FS-Net \cite{chen2021fsnet}  & 14.8 & 3.3 & 4.2  & 17.1 & 21.6 \\
GPV-Pose \cite{di2022gpvpose}  & 15.2 & 3.5 & 4.6  & 17.8 & 22.7 \\
VI-Net \cite{lin2023vinet}  & {20.4} & {8.4} & {10.3}  & {20.5} & {29.1} \\
\textbf { SecondPose (Ours) } & \textbf{24.9} & \textbf{11.0} & \textbf{13.4}  & \textbf{25.3} & \textbf{35.7} \\

\bottomrule
\end{tabular}
\caption{Quantitative comparisons of different methods for category-level 6D object pose estimation on HouseCat6D~\cite{jung2022housecat6d}.}
\label{tab:mainresults_housecat2}
\vspace{-0.5cm}
\end{table*}

We present the categorical results of our experiment on HouseCat6D in Fig.~\ref{fig:category}. Our method exhibits a substantial performance advantage over VI-Net in categories such as box, can, cup, remote, teapot, and shoe. However, in other categories, namely bottle, cutlery, glass, and tube, our method shows a slightly lower performance compared to VI-Net.
We noted a shared characteristic among these categories—items within them typically display either high reflectivity or high transparency. Under optical conditions of this nature, DINOv2 tends to encounter difficulties in extracting meaningful semantic information.

\begin{figure}[t]
    \centering
    \includegraphics[width=\linewidth]{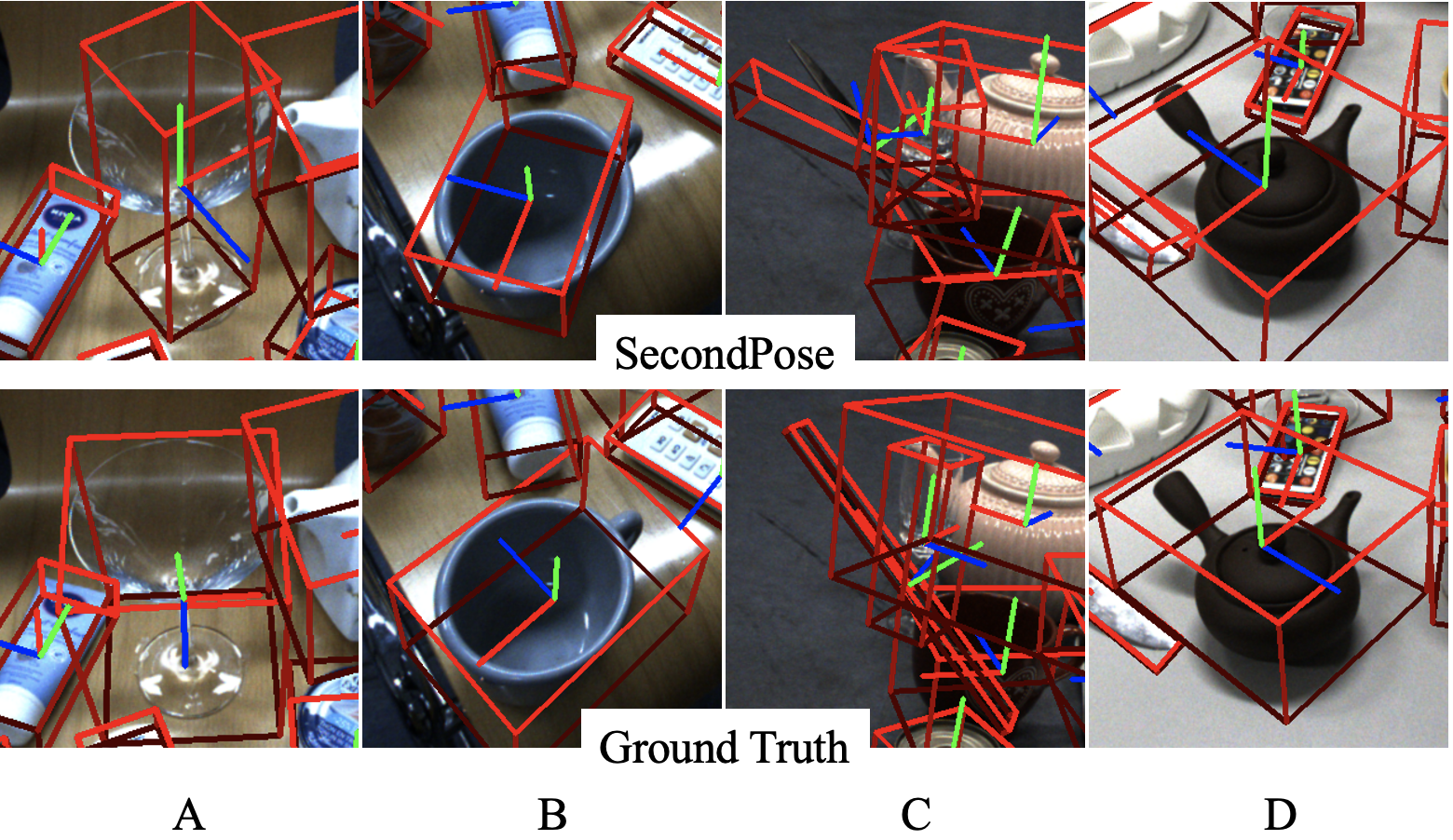}
    \caption{\textbf{Failure cases in HouseCat6D.} We illustrate common failure scenarios on HouseCat6D. (A) depicts instances of transparent items; (B) showcases items with pronounced self-occlusion; (C) the tube represents items with high reflectivity; (D) illustrates failures attributed to atypical shapes.}
    \label{fig:fail_house}
\end{figure}

\begin{figure}[t]
    \centering
    \includegraphics[width=\linewidth]{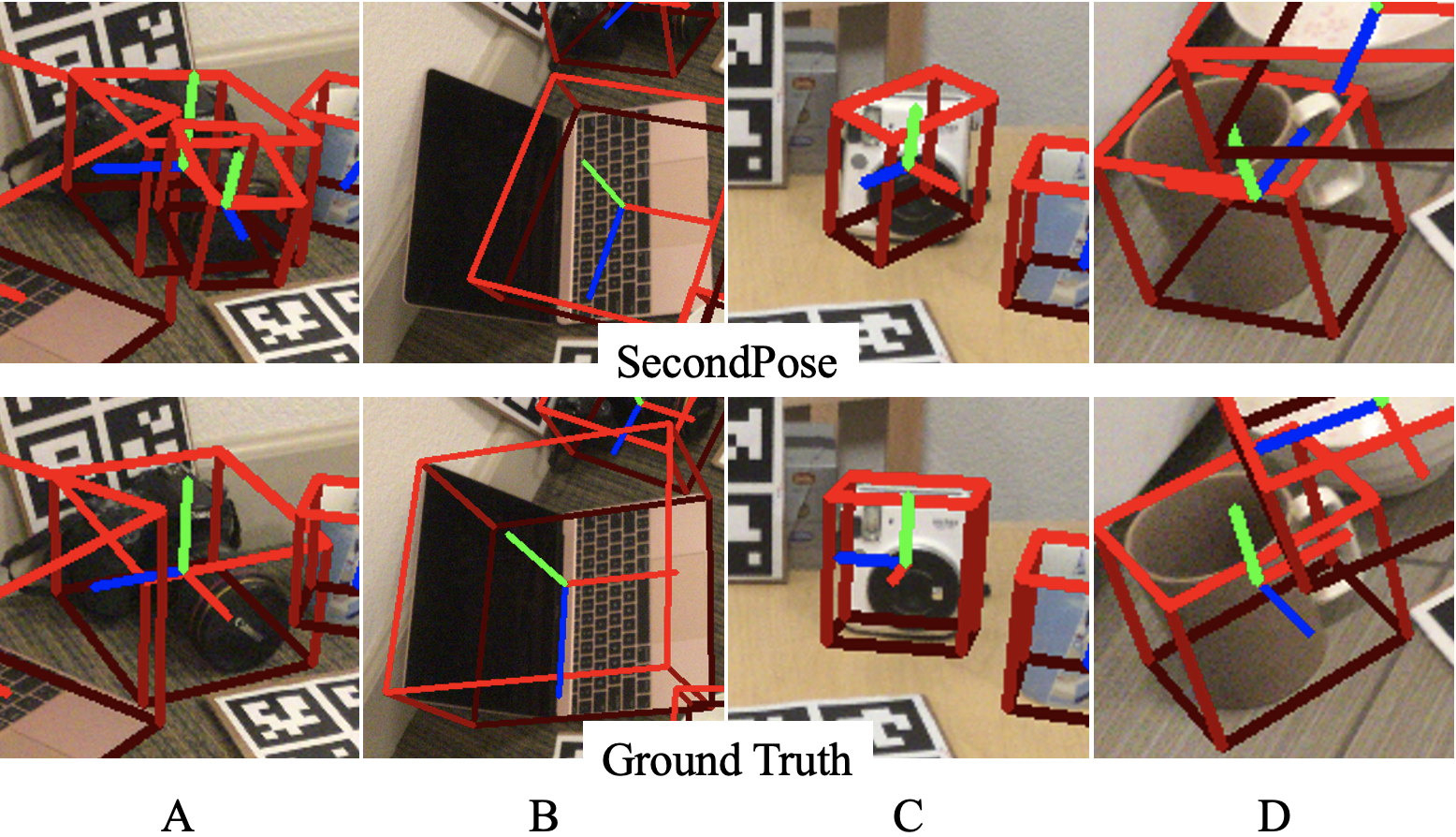}
    \caption{\textbf{Failure cases in REAL275.} We illustrate common failure scenarios on HouseCat6D. (A) represents failure due to wrong instance segmentation; (B)-(D) illustrates failures due to wrong prediction of the y-axis. }
    \label{fig:fail_real}
\end{figure}

\begin{figure*}[t]
    \centering
    \includegraphics[width=\textwidth]{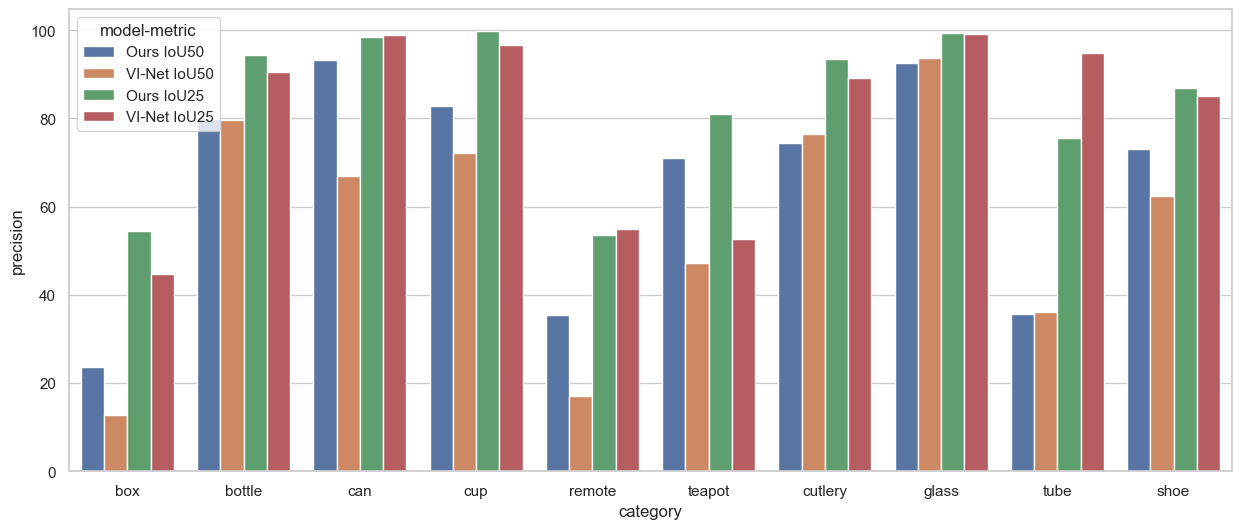}
    \caption{\textbf{Categorical results on HouseCat6D.} We visualize the comparison of our IoU$_{25}$ and IoU$_{50}$ results on HouseCat6D with those of VI-Net. }
    \label{fig:category}
\end{figure*}

\section{Failure Cases and Limitations}
We present typical failure cases in both REAL275 and HouseCat6D.

The failure cases of HouseCat6D are presented in Fig.~\ref{fig:fail_house}. There are four common failure types. (A) highlights instances involving transparent items where DINOv2 struggles to extract meaningful semantic features, leading to poorer performance on transparent items. (B) illustrates a self-occlusion scenario, complicating pose prediction due to obscured essential features like the mug handle, which is crucial for object identification and orientation. In (C), the tube represents items with high reflectivity, a condition often associated with DINOv2 failures. (D) illustrates failures attributed to atypical shapes.

The failure cases of REAL275 are presented in Fig.~\ref{fig:fail_real}. (A) signifies failures arising from false positive detection results. Meanwhile, (B)-(D) illustrate errors related to the wrong orientation prediction of the z-axis, where we observed that on REAL275, our model tends to predict the y-axis accurately.

In summary, there are four primary typical failure scenarios: Firstly, instances where DINOv2 fails to extract meaningful semantic information under specific optical conditions such as high reflectivity or high transparency. Secondly, when severe occlusions are present. Thirdly, when the item displays an atypical shape. Finally, errors are caused by the exclusive parts out of our pipeline, such as the detection frontend.
\end{document}